\PassOptionsToPackage{table}{xcolor}

\documentclass[sigconf]{acmart}

\usepackage{etoolbox}
\undef\Bbbk

\usepackage{amsfonts}
\usepackage{amssymb}

\usepackage{bbm}

\usepackage{algorithm}
\usepackage{subfigure}
\usepackage{subcaption}
\usepackage{multirow}
\usepackage{hyperref}
\usepackage{algpseudocode}

\AtBeginDocument{%
  }

\copyrightyear{2025}
\acmYear{2025}
\setcopyright{acmlicensed}\acmConference[WWW Companion '25]{Companion Proceedings of the ACM Web Conference 2025}{April 28-May 2, 2025}{Sydney, NSW, Australia}
\acmBooktitle{Companion Proceedings of the ACM Web Conference 2025 (WWW Companion '25), April 28-May 2, 2025, Sydney, NSW, Australia} 
\acmDOI{10.1145/3701716.3715226} 
\acmISBN{979-8-4007-1331-6/2025/04}

\begin{document}

\title{GAS: Generative Auto-bidding with Post-training Search}



\author{Yewen Li}
\authornote{Both authors contributed equally to this research.}
\affiliation{%
  \institution{Nanyang Technological University}
  \city{Singapore}
  \country{Singapore}
  }
\email{yewen001@e.ntu.edu.sg}

\author{Shuai Mao}
\authornotemark[1]
\affiliation{%
  \institution{The Chinese University of Hong Kong}
  \city{Hong Kong}
  \country{China}}
\email{smao@mae.cuhk.edu.hk}

\author{Jingtong Gao}
\affiliation{%
  \institution{City University of Hong Kong}
  \city{Hong Kong}
  \country{China}}
\email{jt.g@my.cityu.edu.hk}


\author{Nan Jiang}
\affiliation{%
  \institution{Kuaishou Technology}
  \city{Beijing}
  \country{China}
}
\email{jiangnan07@kuaishou.com}

\author{Yunjian Xu}
\affiliation{%
  \institution{The Chinese University of Hong Kong}
  \city{Hong Kong}
  \country{China}}
\email{yjxu@mae.cuhk.edu.hk}

\author{Qingpeng Cai}
\authornote{Corresponding author.}
\affiliation{%
  \institution{Kuaishou Technology}
  \city{Beijing}
  \country{China}
}
\email{caiqingpeng@kuaishou.com}

\author{Fei Pan}
\affiliation{%
  \institution{Kuaishou Technology}
  \city{Beijing}
  \country{China}
}
\email{panfei05@kuaishou.com}

\author{Peng Jiang}
\affiliation{%
  \institution{Kuaishou Technology}
  \city{Beijing}
  \country{China}
}
\email{jiangpeng@kuaishou.com}

\author{Bo An}
\affiliation{%
  \institution{Nanyang Technological University\\Skywork AI}
  \city{Singapore}
  \country{Singapore}
  }
\email{boan@ntu.edu.sg}

\renewcommand{\shortauthors}{Yewen Li et al.}
\begin{abstract}
Auto-bidding is essential in facilitating online advertising by automatically placing bids on behalf of advertisers.  
Generative auto-bidding, which generates bids based on an adjustable condition using models like transformers and diffusers, has recently emerged as a new trend due to its potential to learn optimal strategies directly from data and adjust flexibly to preferences.
However, generative models suffer from low-quality data leading to a mismatch between the condition, like \textit{return to go}, and true action value, especially in long sequential decision-making.
Besides, the majority preference in the dataset may hinder models' generalization ability on minority advertisers' preferences.
While it is possible to collect high-quality data and retrain multiple models for different preferences, the high cost makes it unaffordable, hindering the advancement of auto-bidding into the era of large foundation models.
To address this, we propose a flexible and practical \textbf{G}enerative \textbf{A}uto-bidding scheme using post-training \textbf{S}earch, termed \textbf{GAS}, to refine a base policy model's output and adapt to various preferences. 
We use weak-to-strong search alignment by training small critics for different preferences and an MCTS-inspired search to refine the model's output. 
Specifically, a novel voting mechanism with transformer-based critics trained with policy indications could enhance search alignment performance. 
Additionally, utilizing the search,
we provide a fine-tuning method for high-frequency preference scenarios considering computational efficiency.
Extensive experiments conducted on the real-world dataset and
online A/B test on the Kuaishou advertising platform demonstrate the
effectiveness of GAS, achieving significant improvements, \textit{e.g.}, 4.60\% increment of target cost.
\href{https://github.com/yewen99/GAS_WWW-25}{\textcolor{cyan}{Code is available here}}.
\end{abstract}

\ccsdesc[500]{Information systems~Computational advertising}


\keywords{Auto-bidding, Generative Model, Search, Preference Alignment}


\maketitle

\vspace{-2mm}
\section{Introduction}
The rapid digitalization of commerce has expanded online advertising platforms' reach, making them essential for engaging audiences and boosting sales \cite{DBLP:conf/wsdm/WangY15, evans2009online}. Due to numerous impression opportunities, manually adjusting bids to optimize costs under budget and KPI constraints is impractical.
To address this, ad platforms now offer auto-bidding services that automate the bidding process using advanced strategies \cite{DBLP:conf/nips/BalseiroDMMZ21, DBLP:conf/www/DengMMZ21, DBLP:conf/kdd/OuCYDLZTY23}. These strategies consider factors from immediate or historical bidding information, such as the distribution of impression opportunities and remaining budgets \cite{DBLP:journals/corr/abs-2210-06107}.
Besides, according to different types of advertisers, the strategy should take their different preferences into consideration.
For example, brand advertisers, who aim for long-term growth and awareness, typically aim to show their ads to as many people as possible under constraints like average cost per impression. However, performance advertisers, who focus on maximizing the value of winning impressions, aim to maximize conversions with cost-per-conversion constraints \cite{uscb}.
To meet these diverse needs, advertising platforms like Google, Facebook, and Alibaba provide tailored multiple bidding strategies for their customers \cite{google2021, facebook2021, alibaba2021}.
Additionally, facing the dynamic advertising environment, strategies must be regularly optimized to stay closely aligned with customers' preferences, thereby achieving long-term commercial benefits \cite{aigb}.

Reinforcement learning \cite{sutton2018reinforcement,pan2020softmax} (RL) has long been a leading method for enhancing auto-bidding strategies by training agents with an advertising simulator or offline advertising logs. 
However, RL approaches are predominantly based on the Markovian decision process (MDP), which assumes that future states are determined solely by the current single step's state and action. 
Recent statistical analyses \cite{aigb} question this assumption in auto-bidding, revealing a strong correlation between the lengths of historical state sequences and subsequent states.
This finding suggests that relying solely on the most recent state can lead to instability in the unpredictable online advertising environment.
Moreover, the preference for the RL strategy is not easy to control. 
USCB \cite{uscb} proposes calculating an optimal solution under multiple constraints using historical data and then training RL to refine the strategy. However, once deployed, the strategy's preference is fixed, limiting interactivity and controllability.
Therefore, this has led to a new trend in developing generative auto-bidding methods based on conditional generative models like transformers and diffusers.
With a vector condition representing the preferences, these methods directly output an action or even a trajectory without MDP assumption. For example, decision transformers \cite{dt} can utilize extensive historical information for decision-making. Diffusers \cite{diffuser} can directly generate a trajectory for planning given a condition. More importantly, generative models provide flexibility in controlling preferences by simply modifying their condition's value during deployment. 
As large foundational models based on generative models have shown remarkable progress in applications like natural language processing and computer vision, exemplified by ChatGPT \cite{ChatGPT} and Stable Diffusion \cite{latent_diffusion}, it is foreseeable that the auto-bidding field will also advance towards the foundation model era. This will involve developing a decision foundation model to learn optimal decision strategies directly from large-scale datasets.

However, there are also two major intrinsic challenges in applying generative auto-bidding methods.
First, the performance of generative bidding methods is significantly influenced by the quality of the dataset, { where the collected conditions, \textit{i.e.}, return to go, cannot reflect the true value of the action.
For instance, a good action $a_t$ followed by a bad future action $a_{i}, i>t$, can result in a low return to go for $a_t$, and vice versa.
Thus, the learned strategy is hard to achieve the optimum due to a mismatch between the condition and the true action value during training.}
Second, realistic bidding tasks often involve varying preferences over time, 
but generative methods are always trained or even biased to imitate the majority preference \cite{llm_bias}, necessitating retraining to adapt to new minority preferences.
However, since the scaling law could also happen in this auto-bidding field, the bidding model based on a foundation model like a transformer would get larger and larger, making it impractical and costly to retrain a set of large decision transformers for varying preferences,  hindering the advancement
of auto-bidding into the era of large foundation models. 
Therefore, a question is raised: ``Could we only use one policy model to efficiently achieve optimum strategy under various preferences?''
To address these issues, we propose a \textbf{G}enerative \textbf{A}uto-bidding scheme using post-training \textbf{S}earch, termed \textbf{GAS}, to refine a single base policy model's output and adapt to various preferences efficiently. 
Instead of revising the condition setting and retraining the policy model to fit various preferences, 
we adopt the weak-to-strong search alignment idea, \textit{i.e.}, refining the large foundation model's output with small models \cite{weak2strong}. 
Specifically, we train a set of small critics to assess the value of different preferences, then use a search method to refine the model's output inspired by Monte Carlo Tree Search (MCTS) \cite{mcts}. 
{ In this scheme, the mismatch between the condition value and the true action value, approximated by the critics using Q-learning, would be alleviated.
This is due to the Q-learning based on a Bellman backup
only requires the current reward that is not affected by the future trajectory.
A large dataset collected by various policies is also beneficial to training a critic as it could assess actions of different quality, alleviating the out-of-domain overestimation issue \cite{iql}.
This scheme could also achieve better alignment with different preferences by refining the action guided by these critics without retraining or fine-tuning the model.}
To summarize, our contribution could be summarized as follows:
\begin{itemize}
    \item This paper proposes a flexible and practical framework using a post-training search method for refining and aligning generative auto-bidding models to various preferences, which unlocks the door of auto-bidding foundation models.
    \item To enhance the accuracy of the search process, we utilize transformer-based critics that are aware of the underlying policy by leveraging historical bidding sequences and introduce a novel voting mechanism to enhance value approximation accuracy during the value backpropagation stage of the search process.
    \item In addition to performing a search during testing time, we provide a fine-tuning method for high-frequency preference scenarios or computational-efficiency-sensitive scenarios.
    \item Extensive experiments conducted on a large real-world dataset and an online A/B test on the Kuaishou advertising platform demonstrate the effectiveness of the proposed generative auto-bidding scheme.
\end{itemize}


\section{Preliminary}
\subsection{Problem Statement}
During a time period, suppose there are \(H\) impression opportunities arriving sequentially and indexed by \(i\). In a bidding platform, advertisers submit bids to compete for each impression opportunity. 
An advertiser wins the impression if its bid $b_i$ is higher than the bids of others. Upon winning, the advertiser incurs a cost of $c_i$, which is typically the highest bid of the other advertisers in a second-price auction setting.
During the period, the goal of an advertiser is to maximize the sum value of winning impressions \(\sum_i o_i v_i\), where \(v_i\) is the value of impression \(i\) and \(o_i\) is the binary indicator of whether the advertiser wins impression \(i\). 
Besides, the budget and multiple KPI constraints \cite{uscb} are critical for an ad advertiser to control
the ad delivery performance.
The budget constraint is considered as \(\sum_i o_i c_i \leq B\), where \(c_i\) is the cost of impression \(i\) and \(B\) is the budget. 
Other KPI constraints are more complicated and can be classified into two categories. The first category is the cost-related (CR) constraints, which restrict the unit cost of certain advertising events, such as CPC and CPA. The second category is the non-cost-related (NCR) constraints, which restrict the average advertising effect, such as CTR and CPI. 
For simplicity, we consider the auto-bidding with cost-related constraints, which have a unified formulation:
$
\frac{\sum_i c_{ij} o_i}{\sum_i p_{ij} o_i} \leq C_j,
$
where \(C_j\) is the upper bound of \(j\)-th constraint provided by the advertiser. \(p_{ij}\) can be any performance indicator, \textit{e.g.}, return or constant. \(c_{ij}\) is the cost of constraint \(j\). Given \(J\) constraints, we have the Multi-Constrained Bidding (MCB):
\begin{equation}
\begin{aligned}
\text{maximize} \quad & \sum_i o_i v_i \\
\text{s.t.} \quad & \sum_i o_i c_i \leq B  \\
& \frac{\sum_i c_{ij} o_i}{\sum_i p_{ij} o_i} \leq C_j, \quad \forall j  \\
& o_i \in \{0, 1\}, \quad \forall i
\label{eq:goal}
\end{aligned}
\end{equation}
A previous study \cite{uscb} has already shown the optimal solution:
\begin{equation}
b_i^* = \lambda_0 v_i + \sum_{j=1}^{J} \lambda_j p_{ij}{C_j},
\label{eq:optimal_bid}
\end{equation}
where \(b_i^*\) is the optimal bid prediction for the impression \(i\). 
\(\lambda_j\), \(j \in \{0, \ldots, J\}\) are the optimal bidding parameters.
However, due to the uncertainty and dynamics of the advertising environment, these optimal bidding parameters are hard or even impossible to calculate directly.
Different types of advertisers may have different preferences expressed by various constraint compositions. 
For example, max-return bidding advertisers only consider the budget constraint and target-CPA bidding advertisers consider both the budget constraint and the CPA constraint.

\subsection{Decision-making Process for Auto-bidding}
As the advertising environment is highly dynamic, the optimal bidding parameters should be adjusted regularly to maximize the total received value over the entire period. This leads to the formulation of the auto-bidding task as a sequential decision-making task.
We consider a standard decision-making setup consisting of an auto-bidding agent interacting with advertising environment ${E}$ in discrete timesteps.
At each timestep $t$ of a bidding period, the agent receives a state $s_t\in\mathcal{S}$ describing the real-time advertising status and then outputs an action $a_t\in\mathcal{A}$ for the final bid.
The advertising environment has an underlying unknown state transition dynamic $\mathcal{T}$.
Under an MDP assumption, the transition dynamic $\mathcal{T}$ could be expressed by $\mathcal{T}:s_t \times a_t \rightarrow s_{t+1}$, \textit{i.e.}, the next state $s_{t+1} \in \mathcal{S}$ is determined by the current state $s_t$ and action $a_t$.
In this case, the agent's policy is expressed as $\pi(a_t|s_t)$.
Without an MDP assumption, the next state could be determined by more factors like the historical trajectory $\tau$.
After transitioning to the next state, the advertising environment would emit a reward $r_t $ representing the value contributed to the objective obtained within the time period $t$.
Repeating this process until the end of a bidding period, one day for instance, the goal of the auto-bidding agent is to maximize the total received value over the entire period as stated in Equation \ref{eq:goal}.
A detailed description of our modeling are listed below:
\begin{itemize}
    \item \textbf{$s_t$}: the state is a collection of information that describes the advertising status from the perspective of a campaign. The information should in principle reflect time, budget consumption and KPI constraints satisfying status, such as left time, left budget, budget consumption speed, current KPI ratio for constraint \(j\) (i.e. \((\Sigma_i c_{ij} o_i / \Sigma_i p_{ij} o_i) / C_j\)), etc.
    \item \(a_t\): adjustment to the bidding parameters \(\lambda_j\), \(j = 0, \ldots, J\), at the time period \(t\),
    and modeled as \((a_t^{\lambda_0}, \ldots, a_t^{\lambda_J})\). 
    \item $r_t$ : at time-step \(t\), let \(\mathcal{C}\) be the candidate impression set between step \(t\) and \(t + 1\), then a typical setting of the reward could be the value contributed to the objective obtained on this \(\mathcal{C}\) within the time period.
    For simplicity, we will omit $\mathcal{C}$ in the following sections.
\end{itemize}


\section{Generative Auto-bidding with Search}
{In this section, we first introduce how an MCTS-inspired post-training search process is developed to refine a base policy model's output action and then, we introduce two practical paradigms of applying this post-training search in auto-bidding.}

\subsection{MCTS-inspired Post-training Search} \label{sec:search}
As the decision transformer is widely used as a backbone in generative decision-making, we also adopt the decision transformer as our backbone for auto-bidding and the policy for generating an action could be formulated as
\begin{align}
    a_t \sim \pi_{dt}= \text{DT}_{\theta}(a|s_{\leq t}, a_{<t}, R_{\leq t}),
\end{align}
where the condition $R_t$ is the \textit{return to go} at time step $t$, \textit{i.e.}, 
\begin{align}
    R_t = \sum_{i=t\sim T}\gamma^{i-t}r(s_i, a_i),
\end{align}
where $\gamma$ is the discounting factor and $r(s_i, a_i)$ is a rewarding function representing the preference, \textit{e.g.}, it could be set as the $o_i v_i$ representing a preference considering only the value. 

\begin{figure*}[htbp]
    \centering
    \subfigure[GAS: Generative Auto-bidding with Post-training Search]{
    \includegraphics[width=0.64\linewidth]{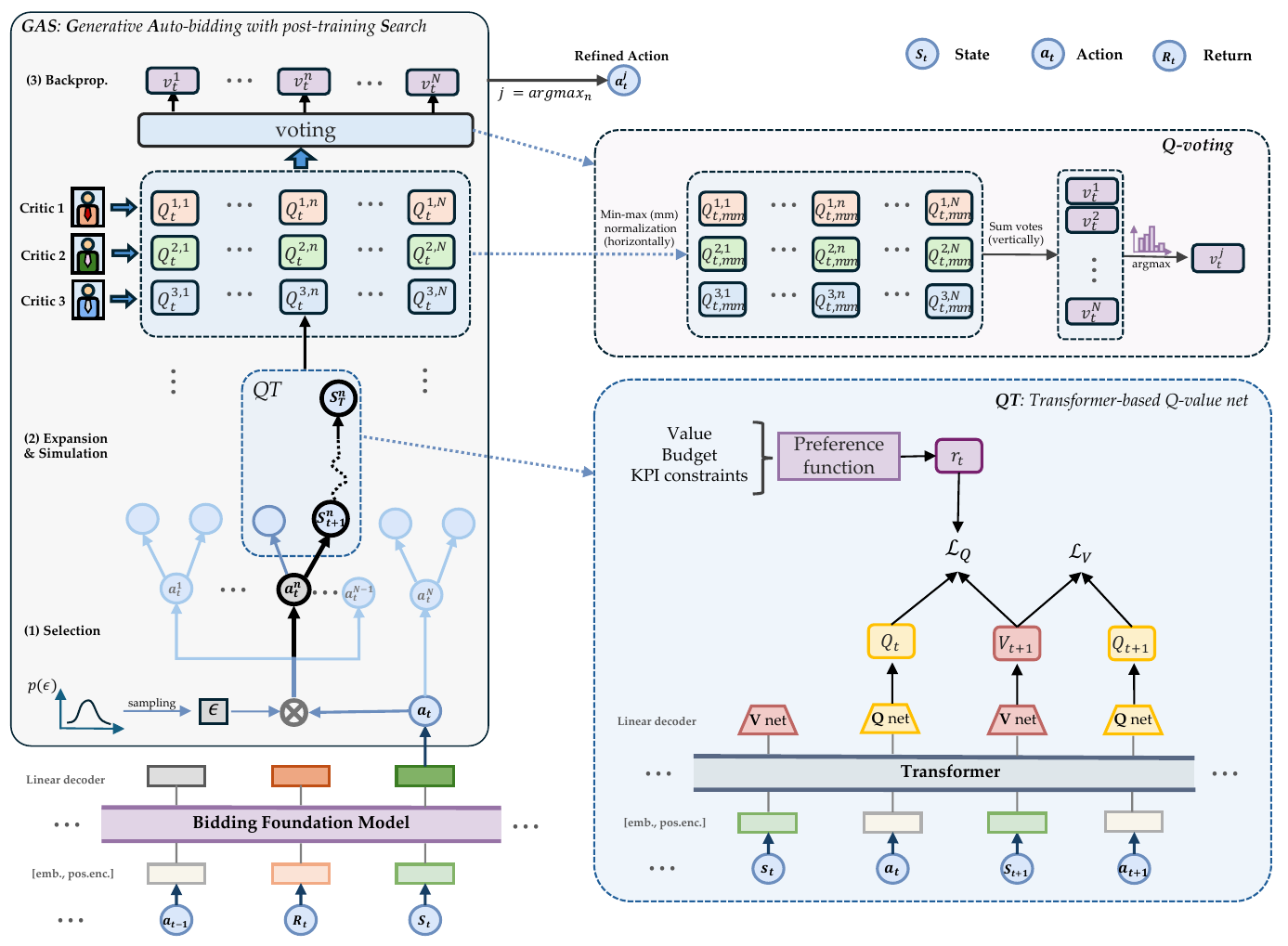}
    \label{fig:framework}
    }
    \hfill
    \subfigure[Online Auto-bidding System]{
    \includegraphics[width=0.33\linewidth]{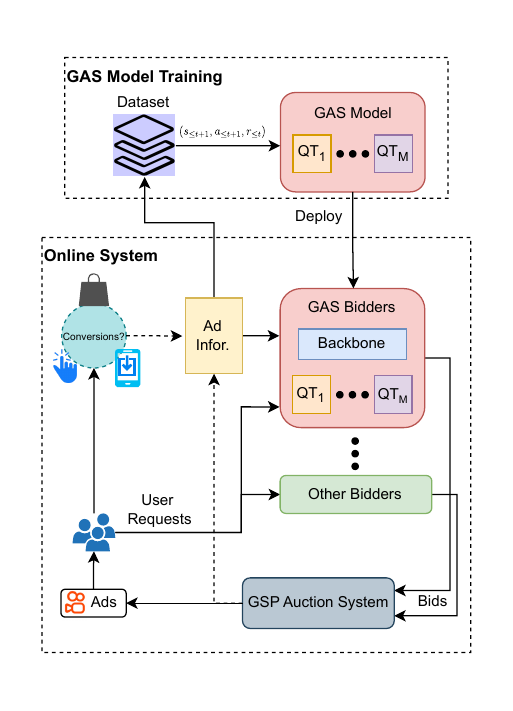}
    \label{fig:online}
    }
    \vspace{-3mm}
    \caption{a) The pipeline of GAS for refining a base action $a_t$ to an action $a_t^j$ better aligned to the preferences represented by the critics, involving three stages: (1) selection by a sampled noisy perturbation, (2) expansion and simulation approximated by the transformer-based Q-value net (QT), and (3) backpropagation through a Q-voting mechanism for better value assessment. b) The online auto-bidding system of Kuaishou including the interaction between the bidding environment and our GAS method.}
    \label{fig:both_images}
\end{figure*}

Our purpose of the search scheme is to find a better action that could align better with the preference like a higher value.
Applying the typical MCTS method to the decision-making process should involve four parts in every time step $t$:  
\begin{itemize}
    \item Selection: start from a root state node $s_t$ and randomly select successive valid child action nodes $a_t^i$ within exploration budget $N$, \textit{i.e.}, $i= 1\sim N$. 
    \item Expansion: unless the child action nodes end the bidding process, create one further child state node $s_{t+1}^i$ with respect to the transition dynamic $s_{t+1}^i \sim \mathcal{T}$. 
    \item Simulation: complete one rollout from node $s_{t+1}^i$ given the policy $\pi(a|s)$ until the end. 
    \item Backpropagation: use the result of the rollout to update value information in the nodes $a_t^i$ given root $s_t$.
\end{itemize}
After these four parts, we could choose a final action $a_t^j$ based on some principles such as balance between exploration and exploitation.
In this work, we choose the action of maximum preference value to execute given a state $s_t$ with uncertainty consideration modeled as a random action selection operation.
However, unlike the GO game where we can simulate all possible moves, this could be impractical in the bidding scenarios as bidding is a typical partially observable MDP (POMDP) task where other advertisers' behaviors are not predictable. 
Therefore, we make modifications to the typical MCTS process by approximating the expansion and simulation process together via learning an enhanced transformer-based Q-value function, without actual simulation in a simulator (which is also invalid). 
Now, we will introduce how we implement \textbf{GAS} in the following three parts, with an illustration in Figure ~\ref{fig:framework}.


\subsubsection{\textbf{Selection}}
Given a decision transformer policy $\text{DT}_\theta$, we generate a base action $a_t^i$ first, and then we perpetuate it by multiplying a random factor uniformly between $90\%$ and $110\%$ to get $N-1$ random actions $\{a_t^i\}_{i=1:N-1}$, expressed as 
\begin{align}
    a_t^i = a_t * \epsilon, \epsilon\sim \mathcal{U}(90\%, 110\%).
\end{align}
We also preserve the initial base action $a_t$ to get the final $N$ action proposals $\{a_t^i\}_{i=1:N} = \{a_t^i\}_{i=1:N-1}\oplus a_t$. Then, the selection process is choosing an action from these action proposals.

\subsubsection{\textbf{Expansion and Simulation}}
As we cannot rollout in a simulator or real environment, we need to approximate this rollout process after taking an action proposal $a_t^i$ given $s_t$. As we actually need only the result of the rollout, we could directly estimate the value of $a_t^i$ given $s_t$ using a Q-value function, expressed as 
\begin{align}
    Q_\phi(s_t,a_t^i;\pi) = r(s_t,a_t^i) + \mathbb{E}_{s_{t+1}\sim \mathcal{T}, a_{t+1}\sim\pi} Q_\phi(s_{t+1}, a_{t+1};\pi).
    \label{eq:Q_pi}
\end{align}
We could employ the IQL \cite{iql} method to the learning of $Q_\phi$, which introduces an extra value net $V_\psi(s)$ to avoid the overestimation issues due to the distributional shift, expressed as
\begin{align}
    \mathcal{L}_{V}(\psi) = \mathbb{E}_{(s,a)\sim \mathcal{D}}[L_2^\tau (Q_{\hat{\phi}}(s,a) - V_\psi(s))],
    \label{eq:v_learning}
\end{align}
where $L_2^\tau(u)=|\tau - \mathbbm{1}(u<0)|u^2$ is an expectile regression loss. The value net $V_\psi(s)$ is used to the Q-value learning:
{\small
\begin{align}
    \mathcal{L}_Q(\phi) = \mathbb{E}_{(s_t,a_t,s_{t+1})\sim\mathcal{D}}[r(s_t,a_t)+\gamma V_\psi(s_{t+1}) - Q_\phi(s_t,a_t))^2].
    \label{eq:q_loss}
\end{align}
}

\noindent As indicated in Equation~\ref{eq:Q_pi}, the Q is coupled with the underlying policy $\pi$ in the latter expectation term.
However, $Q_\phi(s_t,a_t)$ only receives a single state-action pair without policy indication, leading to a value prediction based on the underlying policy $\pi_\beta$ that collected the dataset actually.
This causes a policy gap in the generative auto-bidding, as its policy $\pi_\epsilon$ indicated by various conditions is different from $\pi_\beta$, leading to a biased value approximation.

\noindent\textbf{Rollout via QT.} To address the distributional gap by representing the actual policy $\pi_\epsilon$, we employ the historical trajectory for Q-value learning with a transformer utilizing its sequential modeling ability, termed {\textit{QT}}, \textit{i.e.},
\begin{align}
    Q_\phi^{\pi_\epsilon}(s_t,a_t^i) =Q_\phi(s_t,a_t^i;\pi_\epsilon) = \text{QT}_\phi(s_t,a_t^i;s_{< t}, a_{<t}).
\end{align}
This could be beneficial with a large-scale pertaining set, which contains trajectories collected by various policies because it could empirically use the historical trajectory to predict the future trajectory, \textit{i.e.}, the rollout $\{s_{\leq t}, a_{\leq t}\} \rightarrow \{s_{t+1:T}, a_{t+1:T}\}$.

\noindent After training, the $\text{QT}_\phi(s_t, a_t^i; s_{<t}, a_{<t})$ could return an approximation of the value of the rollout until the end of bidding.

\subsubsection{\textbf{Backpropagation}}
As the overestimation issue of the Q-value function is notorious, an inaccurate value estimation for the rollout could bias the backpropagation to provide an inaccurate value of action node $a_t^i$ given root node $s_t$, leading to the execution of a bad action. To alleviate the overestimate issue, we provide a Q-voting mechanism for value assessment, which is based on a heuristic insight of the consensus.

\noindent\textbf{Value Backpropagation via Q-voting.} Given the success of offline RL, where Q-value is employed to learn a policy improving over the behavior policy that collected the dataset, we could intuitively believe different randomly trained Q-value nets could achieve an agreement in giving the true best action higher value with high probability. Besides, due to the overestimation being an occasional error, there would be no agreement in giving a specific action a higher value. 
Formally, if we independently train $M$ Q-value nets $\{Q_{\phi_k}^{\pi_\epsilon}\}_{k=1:M}$ with different random seeds and we have $N$ action proposals $\{a_t^i\}_{i=1:N}$ with a ground-truth best action $a_t^j$, we model the insight of the consensus by a probability density function $p(a|Q)$ of an action $a$ being the best action given a Q-value net, \textit{i.e.},
\begin{align}
    \textbf{Consensus:  } p(a_t^j|Q_{\phi_k}^{\pi_\epsilon}) > p(a_t^{i\neq j}|Q_{\phi_k}^{\pi_\epsilon}), \forall k\in\{1,...,M\}.
\end{align}
Thus, if we only use a single Q, the final win rate of choosing $a_t^j$ over not choosing it is defined as
\begin{align}
    \mathcal{R}^k := \frac{p(a_t^j|Q_{\phi_k}^{\pi_\epsilon})}{\sum_{i\neq j}p(a_t^i|Q_{\phi_k}^{\pi_\epsilon})}.
\end{align}
\noindent With this consensus, we could apply the majority voting method to increase the win rate $\mathcal{R}^k$. For brevity, we assume $p(a_t^i| Q_{\phi_k}^{\pi_\epsilon})$ is the same for all $k\in\{1,...,M\}$. The probability of choosing an action $a_t^i$ as the final action by majority voting is expressed as
{\small
\begin{align}
    p(a_t^i|\{Q_{\phi_k}^{\pi_\epsilon}\}_{k=1:M})= \sum_{l=\lfloor \frac{M}{2} \rfloor+1}^{M} \binom{M}{l} p(a_t^i|Q_{\phi_k}^{\pi_\epsilon})^l(1-p(a_t^i|Q_{\phi_k}^{\pi_\epsilon}))^{M-l}. \notag
\end{align}
}
Employing the \textit{Condorcet's jury theorem} \cite{boland1989majority}, we could get
\begin{align}
    \mathcal{R}^{1:M} = \frac{p(a_t^i|\{Q_{\phi_k}^{\pi_\epsilon}\}_{k=1:M})}{\sum_{i\neq j}p(a_t^i|\{Q_{\phi_k}^{\pi_\epsilon}\}_{k=1:M})} > \mathcal{R}^k.
\end{align}
To ease reading, let us assume $p(a_t^1|Q_{\phi_k}^{\pi_\epsilon})=0.4$, $p(a_t^2|Q_{\phi_k}^{\pi_\epsilon})=0.3$, and $p(a_t^3|Q_{\phi_k}^{\pi_\epsilon})=0.3$, $\forall k$. Then, we can get $\mathcal{R}^{1:3}=0.81 > \mathcal{R}^{k}=0.67$.

To avoid the useless majority voting outcomes, \textit{i.e.}, there are no actions that gain more votes than all others, we provide a soft majority voting mechanism using Q-value, termed \textbf{\textit{Q-voting}}, implemented by two steps:
\begin{itemize}
    \item step 1: for a $Q_{\phi_k}^{\pi_\epsilon}$, its vote for each action based on min-max normalization over all $a_t^n$ is 
    {\small
    \begin{align}
    v(a_t^i|Q_{\phi_k}^{\pi_\epsilon}) = \frac{Q_{\phi_k}^{\pi_\epsilon}(s_t,a_t^i) - min_n\{Q_{\phi_k}^{\pi_\epsilon}(s_t,a_t^n)\}}{max_n\{Q_{\phi_k}^{\pi_\epsilon}(s_t,a_t^n)\}- min_n\{Q_{\phi_k}^{\pi_\epsilon}(s_t,a_t^n)\}} \in[0,1].
    \end{align}
    }
    \item step 2: the final total votes gained of an action $a_t^i$ is 
    \begin{align}
        v(a_t^i|\{Q_{\phi_k}^{\pi_\epsilon}\}_{k=1:M}) = \sum_{k=1}^M v(a_t^i|Q_{\phi_k}^{\pi_\epsilon}).
    \label{eq:voting_value}
    \end{align}
\end{itemize}

After the above MCTS-inspired search process, we can now get a refined $a_t^n$ that should have higher preference values than $a_t$.


\subsection{Bidding with GAS}
There could be two ways to apply the GAS: \textit{i}) searching in the testing time, and \textit{ii}) using the search to fine-tune the base policy model.
Since both of these approaches need to train a set of critics, we introduce how we train critics representing different preferences.

\noindent\textbf{Preference Representation.} 
The preference is expressed by the setting of the reward function.
We introduce three examples:
\begin{itemize}
    \item Preference that maximizes the winning value under only budget constraint without considering the KPI constraints: $r_t = o_t v_t$, which is actually the Max Return Bidding;
    \item Preference that maximizes a comprehensive performance considering both the winning value and the KPI constraints: $r_t = o_t v_t \cdot \frac{1}{J}\sum_j min\{(\frac{C_j}{c_{tj}o_t/p_{tj}o_t})^\beta, 1\}, \beta>1$;
    \item Preference that prefers more on the KPI constraints by introducing a larger and controllable reweight factor $w$: 
    $r_t = o_t v_t + \frac{w}{J} \sum_j min\{(\frac{C_j}{c_{tj}o_t/p_{tj}o_t})^\beta, 1\}, \beta>1$.
\end{itemize}
Then, we could use these reward functions to train the critics based on Equation \ref{eq:q_loss}.

\subsubsection{\textbf{Inference with Search}}
By repeating the search process introduced in Section \ref{sec:search} at every time step during inference, we can get a testing-time version of our method, termed \textit{GAS-infer}, which could be directly deployed with a base policy model and multiple critics. For a more clear understanding, we provide a detailed procedure in Algorithm \ref{algo:infer}.

\begin{algorithm}[]
\caption{Inference with Search (\textbf{GAS-\textit{infer}})}
\label{algo:infer}
\begin{algorithmic}[1]
\Require Base policy model $\pi_{\theta}$, critics $\{Q_{\phi_k}^{\pi_\epsilon}\}_{k=1}^M$, initial state $s_0$
\Ensure  Action $a^*_t$ at every time step $t$ of the bidding period

\State Initialize $s_t \leftarrow s_0$
\For{each time step $t$}
    \State Generate base action $a_t \sim \pi_{\theta}(a|s_{\leq t}, a_{<t}, R_{\leq t})$
    \State Generate $N-1$ random actions $\{a_{t}^i\}_{i=1}^{N-1}$ by perturbing $a_t$ with a random factor $\epsilon \sim \text{Uniform}(90\%, 110\%)$
    \State Form action proposals $\{a_{t}^i\}_{i=1}^N = \{a_{t}^i\}_{i=1}^{N-1} \cup \{a_t\}$
    \For{each action proposal $a_{t}^i$}
        \State Estimate value $
        v(a_t^i|\{Q_{\phi_k}^{\pi_\epsilon}\}_{k=1}^{M})$ by Q-voting in Eq. \ref{eq:voting_value}
    \EndFor
    \State Select action $a_t^* = \arg\max_{a_{t}^i} v(a_t^i|\{Q_{\phi_k}^{\pi_\epsilon}\}_{k=1}^{M})$
    \State Execute action $a_t^*$ and observe new state $s_{t+1}$
    \State Update $s_t \leftarrow s_{t+1}$
\EndFor
\State \textbf{return} 
\end{algorithmic}
\end{algorithm}

\subsubsection{\textbf{Finetune with Search}}
As the search method has the potential to find better action surrounding a base action, especially when the base action is poor quality or misaligned with the preference, we could use search to enhance the training data and fine-tune the base policy model.
Firstly, given a data point in the dataset $\{s_t, a_t^{gt}\}$, we could do a search for $a_t^{gt}$ and we could get a better action $a_t^p$ aligning better to the preference.
Then, a simple supervised fine-tune (sft) could implemented based on a loss for training $\text{DT}_\theta$:
\begin{align}
    \mathcal{L}_{\text{DT}}^{\text{sft}}(\theta)=mse(a_t, a_t^p).
\end{align}
Although there could be more choices of preference alignment methods, \textit{e.g.}, DPO \cite{dpo} and RLHF \cite{rlhf}, they are typically trajectory-level based on a query dataset and are reported to be unstable, so we leave them as future works.

\section{Experiment}

\subsection{Setup}
\noindent\textbf{Datasets.} Different from the previous dataset preparation procedure that collects private bidding logs from a non-open-source advertising auto-bidding system, we utilize a new public large-scale real-world bidding dataset, termed AuctionNet, released by Alibaba \cite{auctionNet}.
This dataset, to our knowledge the largest available, includes over 500 million records and offers a sparse version for more challenging cases.
{More details are in Appendix \ref{parameter_setting}.}

\noindent\textbf{Evaluation Metrics.}
For simplicity, we adopt three metrics to evaluate the performance: 
\begin{itemize}
    \item  \text{\textbf{Value}}: the total received value during the bidding period, calculated as $\sum_i o_i v_i$;
    \item \text{\text{Exceeding rate}} of the KPI constraints (\textbf{ER}): by introducing a binary indicator $\mathbb{I} (x_j^h, C_j)$ whether the final KPI performance $x_j^n=\Sigma_i c_{ij} o_i / \Sigma_i p_{ij} o_i$ during a period $h$ exceeds the given constraint $C_j$, and assuming we have $H$ periods in total, we have a exceeding rate of the KPI constraints during $H$ periods, defined by
    \begin{align}
        ER = {\frac{1}{H}}{\sum\nolimits_{h=1\sim H} \sum\nolimits_{j=1\sim J}\mathbb{I} (x_j^h, C_j)}.
    \end{align}
    \item \textbf{Score}: by introducing a penalty term 
    \begin{align}
        penaty_j=min\{(\frac{C_j}{\Sigma_i c_{ij} o_i / \Sigma_i p_{ij} o_i})^\beta, 1\}, \beta=2 
    \end{align}
    the score is a balance between the value and the KPI constraint, \textit{i.e.}, 
    \begin{align}
        score = (\sum_i o_i v_i) \times min\{penaty_j\}_{j=1\sim J}
    \end{align}
\end{itemize}


\noindent\textbf{Baselines.} We compare our method to a wide range of baselines involving both RL and generative methods.
For RL, we compare with online RL approach \textbf{USCB} \cite{uscb}, offline approaches \textbf{BCQ} \cite{bcq}, \textbf{CQL} \cite{cql}, and \textbf{IQL} \cite{iql}.
For generative methods, we compare with \textbf{DiffBid} \cite{aigb}, a generative method based on diffusers, and Decision Transformers \textbf{DT} \cite{dt} and its variants \textbf{CDT} \cite{cdt}, a DT-based method that considers a vector of multiple constraints, {\textbf{DT}-\text{score}}: a DT-based method that utilizes the reward function considering both the winning value and the KPI constraints.

\noindent\textbf{Implementation Details.}
The hyperparameters of baselines are set based on the default values referenced in original papers, with further tuning performed to optimize performance.
{ 
For GAS, it has two components: a base policy model and multiple QT networks. 
The base policy model could be any model and we choose the {\textbf{DT}-{score}} with hyperparameters referenced from the official code provided from \cite{xu2024auto}, which a smaller learning rate at $1e^{-5}$ for fine-tune specifically. 
}
For QT network, we employ 6 attention layers with 8 heads for each and set the hidden size as 512 for all layers with 14M parameters in total, which is lightweight.
{ The total training steps are 400k.}
We adopt the AdamW optimizer \cite{adamw} with a learning rate of $1e^{-4}$, and set the batch size as 128.
During training, we utilize the PyTorch framework on two NVIDIA H100 GPUs.
{ For more detailed hyperparameters of QT networks, please refer to Table \ref{gas_hyperparameters} from Appendix \ref{hyper_setting} }.

\subsection{Performance Comparison with Baselines}
In this experiment, we conducted a performance comparison of various baseline methods under different settings, including varying datasets and budget constraints in the MCB bidding. The dataset used for this evaluation is AuctionNet and its sparse version, and the budgets are set at 50\%, 75\%, 100\%, 125\%, and 150\% of the maximum allowable budget.
The results are indicated by the score as a comprehensive assessment of the performance.
Our method is based on the baseline policy model DT-\textit{score}.

The results are presented in Table \ref{tab:main}, which indicates that both GAS-\textit{infer} and GAS-\textit{sft} consistently outperform other methods across all budget settings, achieving the highest scores for all the respective budgets. 
Other methods such as DT, CDT, and DT-\textit{score} also perform well demonstrating the power of generative models compared to the traditional RL methods like IQL.
We found DiffBid does not perform well in this large-scale task, where a possible reason could be the introduced extra challenges in predicting a whole long trajectory and learning an inverse dynamic model, though it could theoretically alleviate the mismatch between return to go and true action value.
Regarding the stability, as shown in Fig. \ref{fig:stable}, GAS also outperforms the base policy model to be more stable.
{GAS-sft does not perform as well as GAS-\textit{infer}, which could be attributed to encoding additional preference information parameterized in the extra introduced critics and performing clearly during testing. In contrast, the GAS-\textit{sft} backbone fuses the actor and critic within the original model, which may cause ambiguity and weak constraints on its critic ability, leading to worse performance.
Note that the Q-voting procedure for value assessment can be executed in parallel, with each Q-voting step taking around 0.1 seconds. This duration is considerably small compared to the interval from $t$ to $t+1$ at 30 minutes.
}
This suggests that the GAS-\textit{infer} method is particularly effective in large-scale auto-bidding tasks.
\begin{table*}[t!]
\caption{Comparison with baselines in different settings in the MCB task. The best results are bolded and the base policy model's results are underlined to demonstrate the performance improvement of our GAS-\textit{infer} method in the last column.}
\vspace{-3mm}
\setlength\tabcolsep{12pt}
\renewcommand{\arraystretch}{1.2}
\resizebox{\linewidth}{!}{
\begin{tabular}{c|c|ccccccccccc}
\hline
\hline
Dataset    & Budget & {\color[HTML]{1F2329} \textbf{USCB}} & {\color[HTML]{1F2329} \textbf{BCQ}} & {\color[HTML]{1F2329} \textbf{CQL}} & {\color[HTML]{1F2329} \textbf{IQL}} & {\color[HTML]{1F2329} \textbf{DiffBid}} & {\color[HTML]{1F2329} \textbf{DT}} & {\color[HTML]{1F2329} \textbf{CDT}} & {\color[HTML]{1F2329} \textbf{DT}-\textit{score}} & \textbf{GAS}-\textit{sft} & \textbf{GAS}-\textit{infer} &\textit{improve} \\ \hline
          & 50\%   &86                                   &190                                  &113                                  &  164                                & 54                                   & 191                                   & 174                                    & \underline{178}                                    & 192                    & \textbf{193}                  &8.42\% \\
          & 75\%   &135                                   &259                                  &139                                  &   232                               & 100                                   & 265                                   & 242                                    & \underline{268}                                    & 273                   &  \textbf{287}                  &7.08\%\\
AuctionNet & 100\%  &157                                   &321                                  &171                                    & 281                                &  152                                  & 329                                   & 326                                    & \underline{334}                                    & 336                    & \textbf{359}                 & 7.48\% \\
          & 125\%  &220                                   &379                                  &201                                  &  355                                &  193                                  & 396                                   & 378                                    & \underline{395}                                    & 398                     &  \textbf{409}                 &3.54\% \\
          & 150\%  &281                                   &429                                  &238                                  &  401                               & 234                                   &  450                                  & 433                                    & \underline{441}                                    & 448                    & \textbf{461}                  &4.53\%\\ \hline
          & 50\%   & 11.5                                 &17.7                                 &12.8                                 &16.5                                 &  9.87                                  &   14.8                                 & 11.2                                    & \underline{17.7}                                    & 18.0                    & \textbf{18.4}                  &3.95\% \\
          & 75\%   &14.9                                 &24.6                                 &16.7                                 &22.1                                 & 15.4                                   & 22.9                                   & 18.0                                    & \underline{25.9}                                    & 26.4                    & \textbf{27.5}                  &6.17\% \\
AuctionNet-sparse & 100\%  &17.5                                  &31.1                                 &22.2                                 &30.0                                 & 19.5                                  & 29.6                                   &  31.2                                   & \underline{33.2}                                    & 34.3                    & \textbf{36.1}                  &8.73\% \\
          & 125\%  &26.7                                  &34.2                                 &28.6                                 &37.1                                 & 25.3                                   & 34.3                                   & 31.7                                    & \underline{39.6}                                    &  39.7                   &  \textbf{40.0}                &1.01\% \\
          & 150\%  &31.3                                  &37.9                                 &35.8                                 &43.1                                 &  30.8                                  & 44.5                                   & 39.1                                    &   \underline{44.7}                                  &  46.4                   &    \textbf{46.5}              &4.02\% \\ \hline
          \hline
\end{tabular}}
\label{tab:main}
\end{table*}

\subsection{Preference Alignment Performance}
To testify whether search with critics of different preferences could improve the performance of a base policy model to specific preference, we conduct preference alignment experiments in Table \ref{tab:preference}.
In this experiment, we evaluated the alignment results of different preference paradigms using two search methods, GAS-\textit{infer} and GAS-\textit{sft}, compared to the baseline DT-\textit{score} method. T
The preferences considered include Score-first, Value-first, and ER-first. 
Table 2 presents the results, showing that GAS-\textit{infer} and GAS-\textit{sft} consistently provide better alignment performance across different preferences compared to the base policy model, supporting the effectiveness of preference alignment by the search methods.

\begin{table*}[]
\caption{Alignment performance with different preferences by two paradigms of search compared to base DT-\textit{score}. 
}
\vspace{-3mm}
\centering
\setlength\tabcolsep{10pt}
\renewcommand{\arraystretch}{1.}
\resizebox{0.8\linewidth}{!}{
\begin{tabular}{c|c|c|cc|cc|cc}
\hline
\hline
Dataset   & Preference & \textbf{Base}      & \multicolumn{2}{c|}{\textbf{Score-first}}                    & \multicolumn{2}{c|}{\textbf{Value-first}}                   & \multicolumn{2}{c}{\textbf{ER-first}}                       \\
          &       & \textbf{DT}-\textit{score} & \textbf{GAS}-\textit{infer} & \textbf{GAS}-\textit{sft}                  & \textbf{GAS}-\textit{infer} & \textbf{GAS}-\textit{sft}                 & \textbf{GAS}-\textit{infer} & \textbf{GAS}-\textit{sft}                  \\ \hline
          & Score $\uparrow$      &\cellcolor[HTML]{EBEFF8} 334       & \cellcolor[HTML]{EBEFF8}{359} & \cellcolor[HTML]{EBEFF8}336 & 358                         & 335                          & 348                         & 346                         \\
AuctionNet & Value $\uparrow$    & \cellcolor[HTML]{D3E1EE} 373      &  391                       & 339                         & \cellcolor[HTML]{D3E1EE}{402} & 375\cellcolor[HTML]{D3E1EE} & 370                          &  359                        \\
          & ER $\downarrow$       &\cellcolor[HTML]{C0D0E7} 37.8\%       & 27.0\%                         & 10.4\%                         & 29.1\%                         & 43.7\%                         & \cellcolor[HTML]{C0D0E7}{23.5\%} & \cellcolor[HTML]{C0D0E7}{16.7\%} \\ \hline
          & Score $\uparrow$     & \cellcolor[HTML]{EBEFF8} 33.2      & \cellcolor[HTML]{EBEFF8}36.1 & \cellcolor[HTML]{EBEFF8} 34.3 &  35.9                        & 34.1                         & 34.5                         & 33.5                        \\
AuctionNet-sparse & Value $\uparrow$    &\cellcolor[HTML]{D3E1EE} 36.8      & 38.9                         & 36.9                         & \cellcolor[HTML]{D3E1EE}39.2 & \cellcolor[HTML]{D3E1EE}37.8 & 36.4                        &  36.3                        \\
          & ER $\downarrow$        & \cellcolor[HTML]{C0D0E7} 27.0\%      & 29.1\%                         & 25.6\%                         & 31.3\%                         &  33.3\%                        & \cellcolor[HTML]{C0D0E7} 20.8\% & \cellcolor[HTML]{C0D0E7}22.9\% \\ \hline \hline
\end{tabular}}
\label{tab:preference}
\end{table*}

\subsection{Ablation Study}

\begin{figure}[htbp]
    \centering
    \subfigure[Searching Budget]{
    \includegraphics[width=0.45\linewidth]{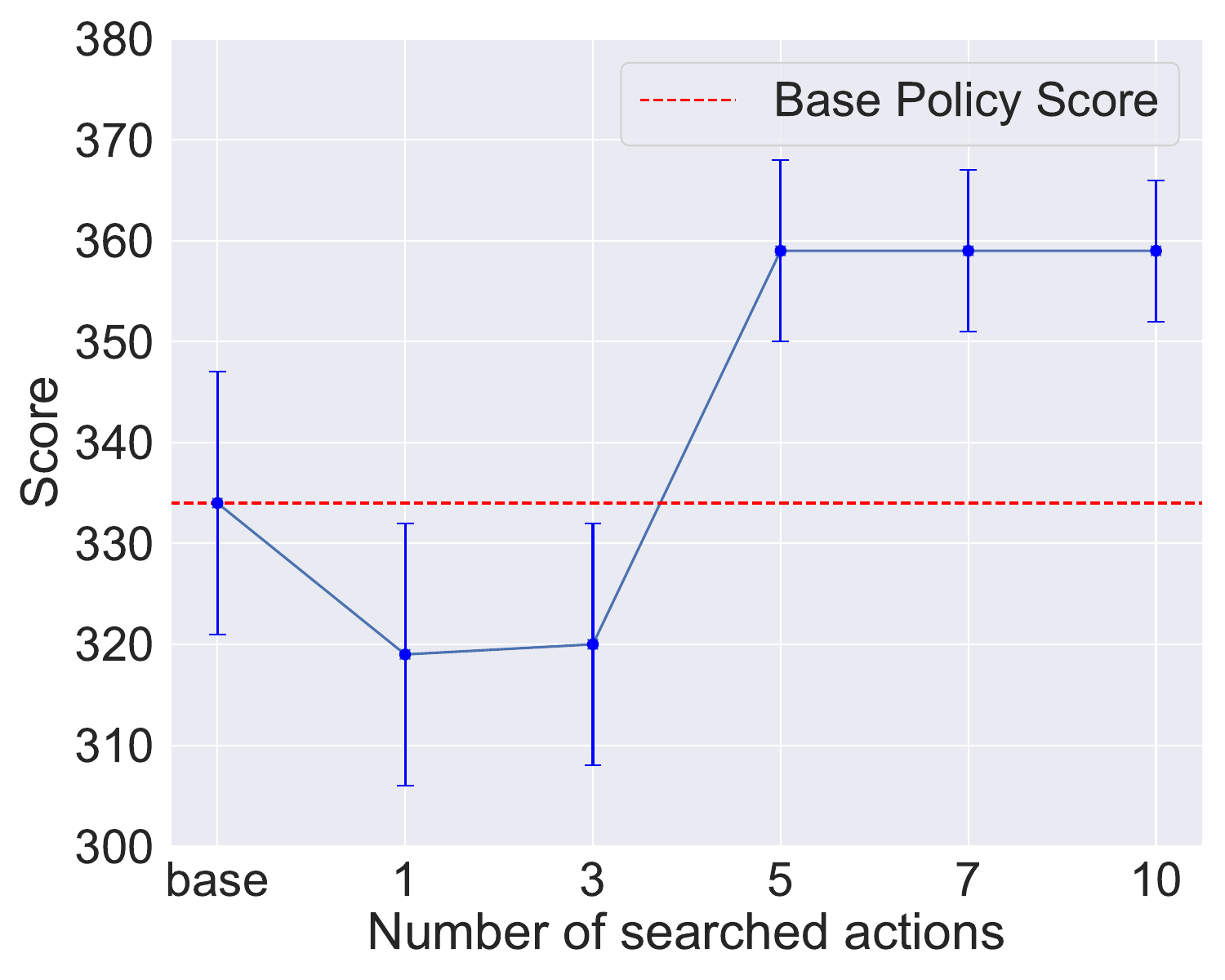}
    \label{fig:num_of_a}
    }
    \hfill
    \subfigure[Number of Critics]{
    \includegraphics[width=0.45\linewidth]{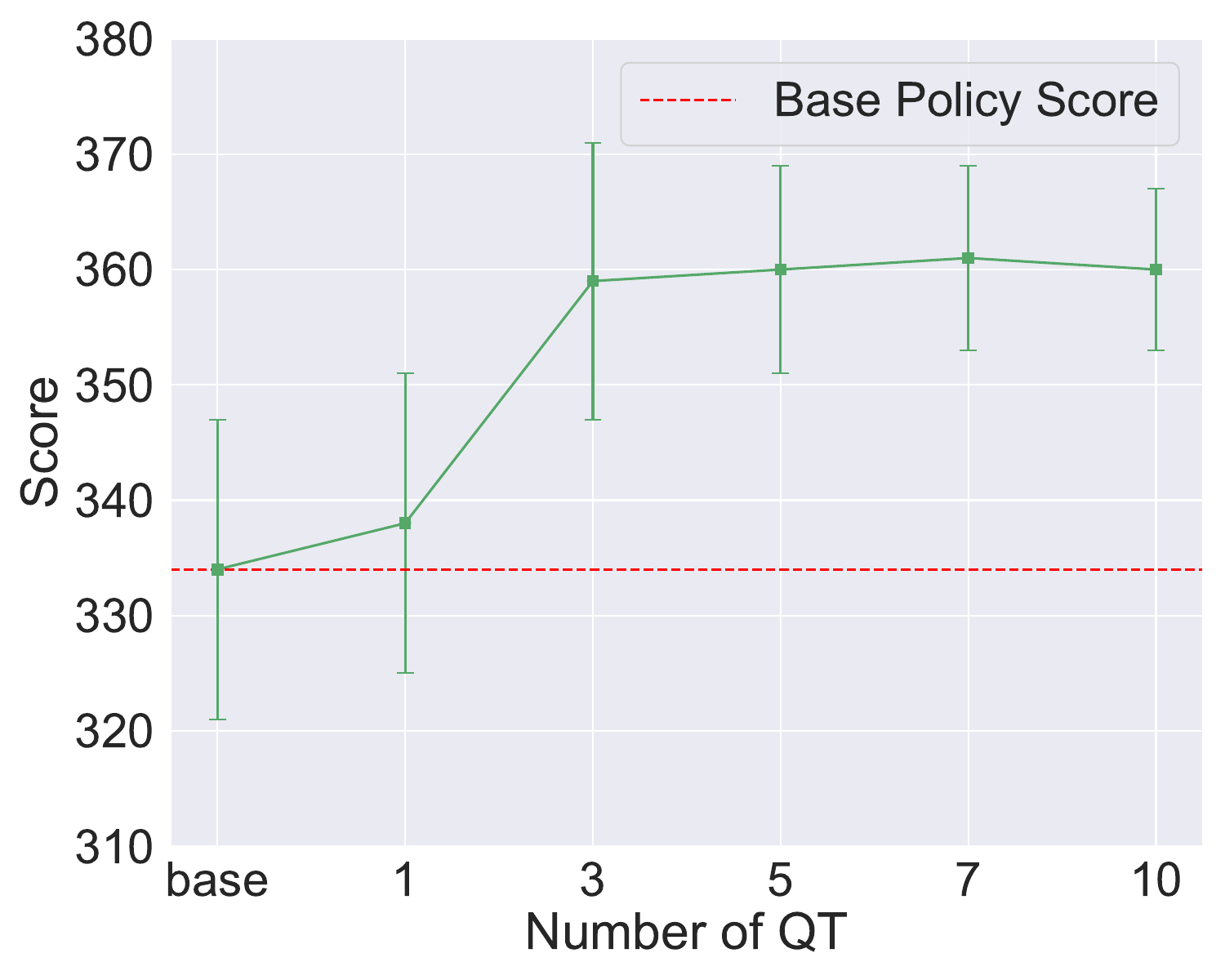}
    \label{fig:num_of_qt}
    }\\
    \subfigure[Range of Search]{
    \includegraphics[width=0.45\linewidth]{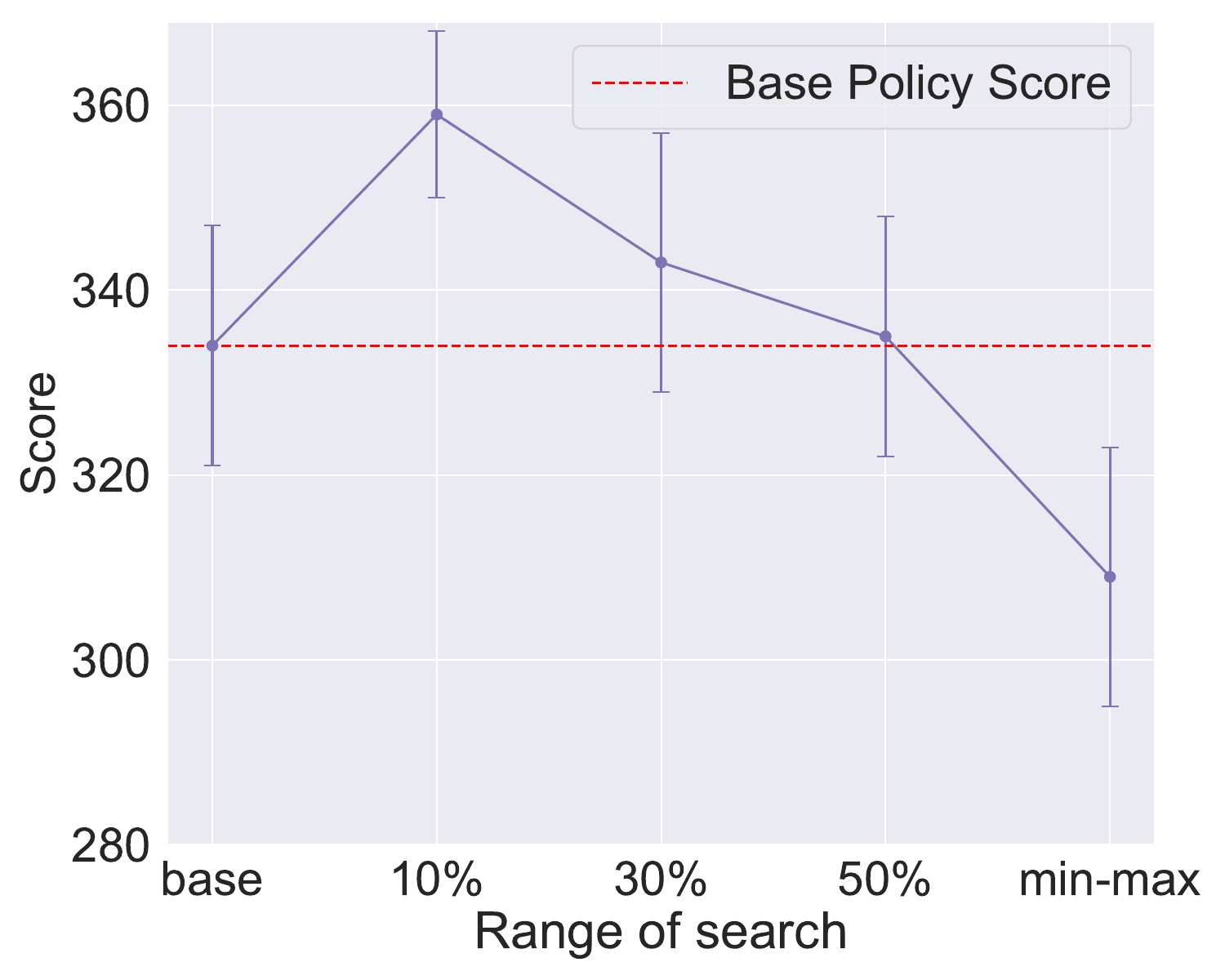}
    \label{fig:range_of_search}
    }
    \hfill
    \subfigure[Stability]{
    \includegraphics[width=0.45\linewidth]{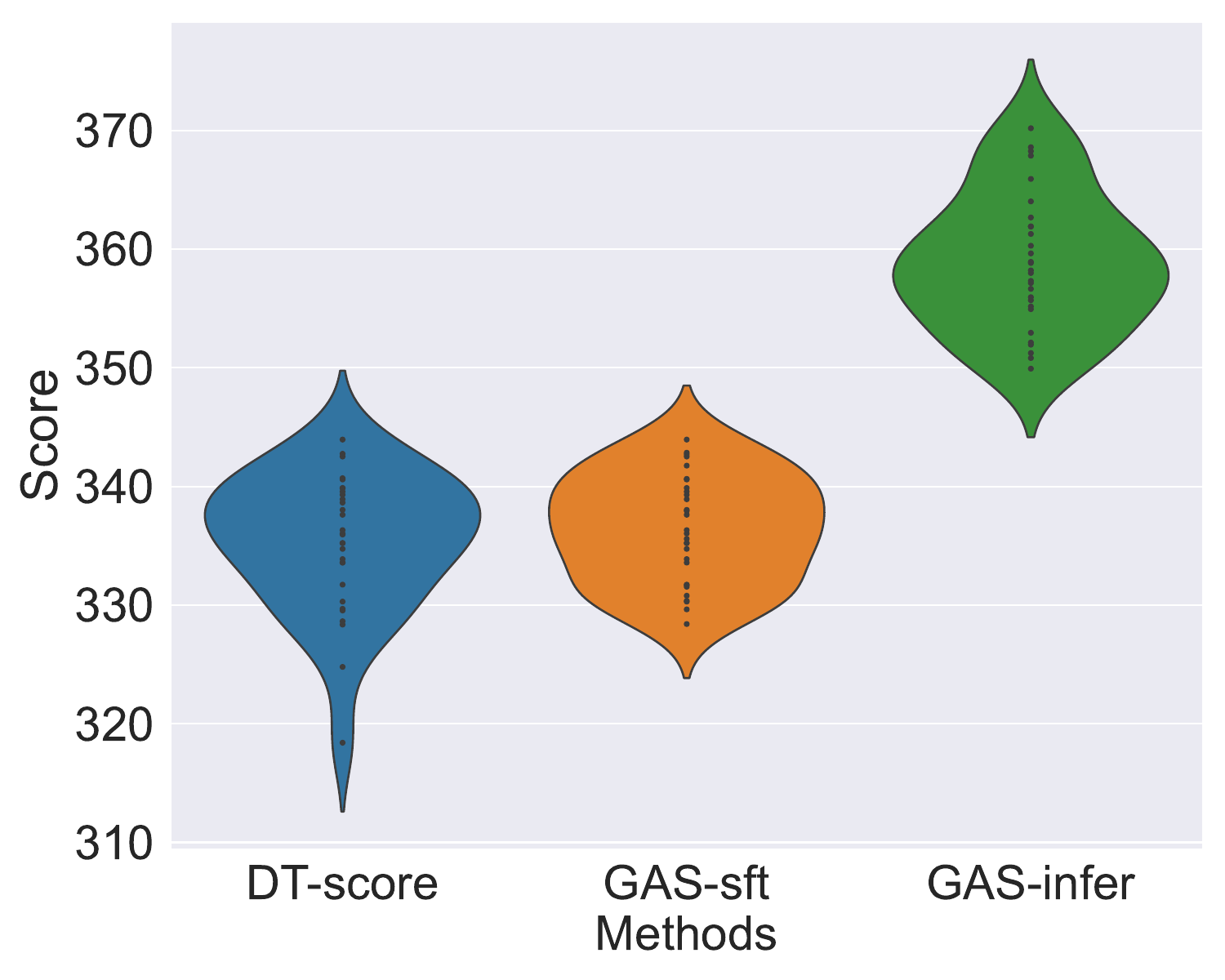}
    \label{fig:stable}
    }
    \vspace{-3mm}
    \caption{\textbf{a-c)} Results for various ablation studies regarding the search setting; \textbf{d)} Comparison of the stability.}
    \label{fig:ablation}
\end{figure}

\noindent{\textbf{Range of Search.}}
The search range directly impacts the balance between exploration (trying out new actions to discover their potential) and exploitation (focusing on actions known to yield good results). A smaller search range means the algorithm will explore fewer actions, potentially missing out on some good actions but focusing more on refining the evaluation of the explored actions. A larger search range means the algorithm will explore more actions, increasing the chances of finding the best possible actions but at the cost of spending less time on each action, which might lead to less accurate evaluations.
An extreme case could be to search the whole action space, while it could be low-efficiency.
Considering the computational efficiency, we randomly search 5 actions within a search range and the ablation study results for different search ranges' results are shown in Fig. \ref{fig:range_of_search}.
The experimental results suggest that for the AuctionNet dataset, a 10\% search range is better, providing the highest performance. Increasing the search range beyond 10\% leads to diminishing returns, and the min-max approach is the least effective due to the limited action budget. 
More importantly, as the optimal performance is achieved with a small search range of around 10\%, it indicates that a small refinement based on the base policy model could be good enough, leaving more space for a more advanced value assessment method.



\noindent{\textbf{Number of Critics.}}
As stated in the Q-voting mechanism, more critics would increase the accuracy of the value assessment empirically, so it would be practical to see how many critics are enough. 
Therefore, we conduct an ablation study on the number of critics. The search range is fixed as $\pm 10\%$ and five random actions are selected for value assessment. As shown in Fig. \ref{fig:num_of_qt}, the experimental results suggest that increasing the number of QT from one to more improves performance significantly, up to an optimal point of seven critics. Beyond this point, further increases from three critics do not yield significant improvements, suggesting that three critics are enough for the current auto-bidding task.



\noindent{\textbf{Searching Budget.}}
Sampling more actions from a search range could increase the possibility of finding the optimal action, 
we conduct an ablation study to investigate it.
Note that, in this investigation, we do not use the original base action.
As shown in Fig. \ref{fig:num_of_a}, we find that the performance can be efficiently enhanced by increasing the number of actions from one to five. However, beyond five actions, the performance remains unchanged.
This suggests that with a higher search budget, we are more likely to find actions that are optimal or near-optimal.


\noindent{\textbf{Effectiveness of QT.}}
As the rollout process is approximated by the Q-value function, the accuracy of the Q-value function in predicting the expected return with the underlying policy becomes essential. Without historical trajectory information, predicting the rollout process given a state $s_t$ and action $a_t$ is unclear and highly random. This insight is supported by our experimental results in Table \ref{tab:search}, where the performance of QT with historical trajectory as policy representation achieves much higher performance than the Q-value function based on a vanilla state-action pair.
We also compare QT to various methods without a Q-value net to testify to the effectiveness of MCTS search, including greedy search, and random/mean selection.
Greedy search selects the top action with the highest predicted immediate reward, 
random selection randomly selects an action, and mean selection takes the average value of 5 random actions.
As shown in Table \ref{tab:search}, our method significantly outperforms other methods.

\vspace{-3mm}
\begin{table}[h!]
\caption{Effectiveness of QT.}
\vspace{-3mm}
\scalebox{.8}{
\begin{tabular}{l|cccccc}
\hline
AuctionNet        & DT-\textit{score} & Greedy &Mean &Random &GAS w/o QT & GAS-\textit{infer} \\ \hline
Score & 334  & 228  & 312 & 319 &292 & 359  \\ \hline
\end{tabular}
}
\label{tab:search}
\end{table}
\vspace{-3mm}

\vspace{-3mm}
\section{Live Experiment}

To verify the effectiveness of our proposed GAS, we have deployed it on Kuaishou’s advertising system, see Figure \ref{fig:online}. The deployment scenario is the Multiple Constraint Bidding(MCB). Under the MCB setting, the advertisers set the budget with or without the CPA/ROI constraint, and the bidding strategy aims to get as many conversions as possible under these constraints.
Due to the limited online testing resources and the potential impact on the advertisers' value, we only compared GAS-\textit{infer} with the baseline model, DT, which is currently in production. The experimental setup is as follows:
\begin{itemize}
    \item \textbf{State}: Budget, cost, time-based budget allocation, time-based cost speed, predicted conversion rates, real CPA or ROI states, etc.
    \item \textbf{Action}: Adjustment to the last time bidding coefficient, \(\lambda_t = \lambda_{t-1}+a_t\), \(\lambda_t\) is the bidding coefficients in Eq. \ref{eq:optimal_bid}.
    \item \textbf{Post-training Search}: The critic is trained with the sum value of the winning impressions, and the search is conducted under the value-first setting. The action range for the search is still sampled within ±10\% of the base action, with 5 points being searched.
\end{itemize}

Our online A/B testing is conducted for six full days. For each MCB campaigns, 25\% budget and traffic are allocated to the baseline bidding model and GAS. The results are shown in Table \ref{tab:online}. The experiment results show that GAS improved impressions by +1.65\%, cost by +0.94\%, target cost by 4.60\%, and overall ROI by +3.62\%. All metrics showed significant improvements.

\vspace{-3mm}
\begin{table}[h!]
\caption{Live experiment: Online A/B test result.}
\vspace{-3mm}
\begin{tabular}{l|cccccc}
\hline
                 & Impression & Cost & Target Cost & ROI\\ \hline
GAS improve      &  +1.65\%  & +0.94\%    &   +4.60\%   &  +3.62\%   \\ \hline
\end{tabular}
\label{tab:online}
\end{table}
\vspace{-3mm}


\section{Related Work}

\noindent\textbf{Auto-bidding and Offline Reinforcement Learning.}
Initially, classic control methods like PID \cite{pid} were used to optimize budget spending using predefined curves, while OnlineLP \cite{onlineLP} adjusted bids based on traffic predictions. 
With increasing complexity of the online bidding environments, RL algorithms such as USCB \cite{uscb}, SORL \cite{sorl}, and MAAB \cite{maab} became crucial for decision-making. Especially offline RL methods, which learn policies from existing datasets without online interaction, have shown significant success. Notable methods include BCQ \cite{bcq}, which narrows the action space to guide the agent towards on-policy behavior; CQL \cite{cql}, which regularizes Q-values for conservative estimates; and IQL \cite{iql}, which enables multi-step dynamic programming updates without querying Q values of out-of-sample actions during training. However, these methods are constrained by the MDP assumption, whereas generative models show greater potential for auto-bidding.




\noindent\textbf{Generative Models.}
Generative models learn the data distribution of a given dataset or establish conditional probability distributions between variables.
Deep generative models like VAEs \cite{vae}, Flows \cite{gflownet}, and GANs \cite{gan} have ushered in a new era of mastering complex data distributions by representing high-dimensional data in a controllable Gaussian form.
Recently, the decision transformer (DT) \cite{dt,liu2024sequential,gao2025future} has been employed to model complex decision-making distribution as an auto-regressive model.
Diffusion models \cite{diffusion} generate samples based on a condition by iterative denoising via a reverse process.
For auto-bidding, DiffBid \cite{aigb} employs a conditional diffusion model to 
generate a long trajectory based on a condition like return and then uses an inverse dynamic model to learn how to map the current state to the predicted next state.

\noindent\textbf{Preference Alignment.}
There are two major ways to finetune a base model to specific preferences, including supervised fine-tuning (SFT) and preference optimization from human feedback.
For SFT \cite{sft}, it constructs high-quality trajectories on specific domains to align the base model to these preferences.
For the latter, it needs to learn a preference model.
Reinforcement learning from human feedback (RLHF) \cite{rlhf}, as a pioneering approach, uses a preference model and proximal policy optimization (PPO) \cite{ppo} to further refine the model. 
Direct preference optimization (DPO) \cite{dpo} and its variants \cite{ipo,cpo,simpo,kto,rdpo} then proposes to directly fine-tune a model using preference query data, bypassing the reward modeling phase and the RL optimization process.
Recently, some post-training methods have been proposed to align the preference even without finetuning by using reward models only \cite{khanov2024args, li2024cascade}.
\section{Conclusion}

This paper presents a flexible and practical framework for generative auto-bidding, termed GAS, which leverages post-training search methods to refine generative auto-bidding policy models with various advertiser preferences. The proposed method utilizes transformer-based critics and a voting mechanism to enhance value approximation accuracy. 
This approach could open new avenues for the application of foundation models in auto-bidding by addressing the challenge of adapting a single model to diverse preferences without the need for multiple costly training processes.

However, there are several \textbf{limitations} in this study. Firstly, the simplifications made in the MCTS process, such as approximating the expansion and simulation steps, may not fully capture the full complexities of real-world bidding scenarios. 
Secondly, the fine-tune version of GAS is more computationally efficient than the inference version, but its performance is still limited, which needs more advanced and effective fine-tune methods. 

\begin{acks}
This research is supported by the National Research Foundation, Singapore, under its Competitive Research Programme (Grant No. NRF-CRP23-2019-0006). Additionally, Shuai Mao and Yunjian Xu are supported in part by the General Research Fund (GRF) project 14200720 of the Hong Kong University Grants Committee, and the National Natural Science Foundation of China (NSFC) Project 62073273.
\end{acks}



\balance
\bibliographystyle{ACM-Reference-Format}
\bibliography{sample-base}

\newpage
\clearpage
\appendix

\section{Appendix}
\label{parameter_setting}

\subsection{Datset Details}

The datasets consisting of AuctionNet and AuctionNet-sparse, where AuctionNet-sparse is a sparse version of AuctionNet with less conversions .
Each dataset consists of 21 advertising delivery periods, with each period containing approximately 500000 impression opportunities, divided into 48 intervals. Detailed parameters are shown in Table \ref{dataset_setting}.
Each advertiser will bid for all impressions.

\begin{table}[h!]
  \centering
  \caption{The Parameters of AuctionNet and AuctionNet-sparse.}
  \scalebox{0.9}{
    \begin{tabular}{ccc}
    \toprule
    Params & AuctionNet & AuctionNet-Sparse \\
    \midrule
    Trajectories & 479,376 & 479,376 \\
    Delivery Periods & 9,987 & 9,987 \\
    Time steps in a trajectory & 48    & 48 \\
    State dimension & 16    & 16 \\
    Action dimension & 1     & 1 \\
    Return-To-Go Dimension & 1     & 1 \\
    Action range & [0, 493] & [0, 589] \\
    Impression's value range & [0, 1] & [0, 1] \\
    CPA range & [6, 12] & [60, 130] \\
    Total conversion range & [0, 1512] & [0, 57] \\
    \bottomrule
    \end{tabular}%
    }
  \label{dataset_setting}
\end{table}%

Each dataset contains over 500 million records, with each recording information for multiple advertisers across various time steps and multiple periods. 
In detail, the state includes the following information:
\begin{itemize}
     \item time\_left: The remaining time steps left in the current advertising period.
     
     \item budget\_left: The remaining budget that the advertiser has available to spend in the current advertising period. 
     
     \item historical\_bid\_mean: The average values of bids made by the advertiser over past time steps.
    
     \item last\_three\_bid\_mean: The average values of bids over the last three time steps. 
    
     \item historical\_LeastWinningCost\_mean: The average of the least cost required to win an impression over previous time steps.
    
     \item historical\_pValues\_mean: The average of historical p-values over past time steps.
    
     \item historical\_conversion\_mean: The average number of conversions (e.g., sales, clicks, etc.) the advertiser achieved in previous time steps. 
    
     \item historical\_xi\_mean: The average winning status of advertisers in impression opportunities, where 1 represents winning and 0 represents not winning.
    
     \item last\_three\_LeastWinningCost\_mean: The average of the least winning costs over the last three time steps. 
    
     \item last\_three\_pValues\_mean: The average of conversion probability of advertising exposure to users over the last three time steps.
    
     \item last\_three\_conversion\_mean: The average number of conversions over the last three time steps.

     \item last\_three\_xi\_mean: The average winning status of advertisers over the last three time steps.
    
     \item current\_pValues\_mean: The mean of p-values at the current time step. 
    
     \item current\_pv\_num: The number of impressions served at the current time step
    
     \item last\_three\_pv\_num\_total: The total number of impressions served over the last three time steps. 
     
     \item historical\_pv\_num\_total: The total number of impressions served over past time steps. 
\end{itemize}

{ We conduct experiments in a simulated experimental environment resembling real advertising system, provided by Alibaba \cite{xu2024auto}.
During evaluation,
an episode, also referred to as an advertising delivery period, is a day divided into 48 intervals, with each interval lasting 30 minutes.
Each episode contains approximately 500000 impression opportunities that arrive sequentially.
An advertising delivery period involves 48 advertisers from diverse categories, with varying budgets and CPAs.
Each advertiser bids for all impression opportunities during each period.
In each evaluation, our well-trained model represents a designated advertiser in bidding given specific budget and CPA. In order to comprehensively evaluate the performance of the model under different advertisers, we will use different advertiser configurations and advertising periods to evaluate this model multiple times in the simulated environment, and average the results as the evaluation score.
}

\subsection{HYPERPARAMETERS SETTING}
We list the hyperparameter details in Table \ref{gas_hyperparameters} for reproduction.

\label{hyper_setting}
\begin{table}[h!]
\caption{The detailed hyperparameters of QT networks}
\scalebox{0.9}{
\begin{tabular}{c|c}
\hline
Hyperparameters     & Value  \\ \hline
Batch size          & 128    \\
Number of steps     & 400000 \\
Sequence length     & 20     \\
Learning rate       & 1e-4   \\
Number of attention layers       & 6   \\
Number of heads       & 8   \\
Optimizer           & AdamW  \\
Optimizer eps       & 1e-8   \\
Weight decay        & 1e-2   \\
Scale               & 2000   \\
Episode length      & 48     \\
Hidden size         & 512    \\
Activation function & ReLU   \\
Gamma               & 0.99   \\
Tau                 & 0.01   \\
Expectile           & 0.7    \\ \hline
\end{tabular}}
\label{gas_hyperparameters}
\end{table}

\end{document}